\documentclass[letterpaper]{article} 
\usepackage{aaai2026}
\usepackage{times}  
\usepackage{helvet} 
\usepackage{courier}
\usepackage[hyphens]{url}
\usepackage{graphicx}
\usepackage{natbib} 
\usepackage{caption} 
\usepackage{graphicx}
\usepackage{amsmath} 
\usepackage{amssymb} 
\usepackage{multirow}    
\usepackage{makecell} 
\usepackage{tabularx}  
\usepackage{array} 
\usepackage{xcolor} 
\usepackage{booktabs}
\usepackage{algorithm}
\usepackage{algorithmic}

\usepackage{newfloat}
\usepackage{listings}
\DeclareCaptionStyle{ruled}{labelfont=normalfont,labelsep=colon,strut=off} 
\floatstyle{ruled}
\newfloat{listing}{tb}{lst}{}
\floatname{listing}{Listing}

\title{CompEvent: Complex-valued Event-RGB Fusion for \\Low-light Video Enhancement and Deblurring}

\author{
    Mingchen Zhong\textsuperscript{\rm 1},
    Xin Lu\textsuperscript{\rm 1}, Dong Li\textsuperscript{\rm 1,}$^\dagger$\\
    Senyan Xu\textsuperscript{\rm 1}, Ruixuan Jiang\textsuperscript{\rm 1}, Xueyang Fu\textsuperscript{\rm 1}, Baocai Yin\textsuperscript{\rm 2,}$^\dagger$
}
\affiliations{
    \textsuperscript{\rm 1}University of Science and Technology of China \\
    \textsuperscript{\rm 2}iFlytek Research, iFlytek Co., Ltd.\\

    \{zmcsa24010071, luxion, dongli6, syxu, sa25010021\}@mail.ustc.edu.cn, xyfu@ustc.edu.cn, bcyin@iflytek.com
}

\usepackage{bibentry}

\begin{document}

\maketitle

\renewcommand{\thefootnote}{$^\dagger$} 
\footnotetext[1]{Corresponding author.}

\begin{abstract}

Low-light video deblurring poses significant challenges in applications like nighttime surveillance and autonomous driving due to dim lighting and long exposures. While event cameras offer potential solutions with superior low-light sensitivity and high temporal resolution, existing fusion methods typically employ staged strategies, limiting their effectiveness against combined low-light and motion blur degradations. To overcome this, we propose CompEvent, a complex neural network framework enabling holistic full-process fusion of event data and RGB frames for enhanced joint restoration. CompEvent features two core components: 1) Complex Temporal Alignment GRU, which utilizes complex-valued convolutions and processes video and event streams iteratively via GRU to achieve temporal alignment and continuous fusion; and 2) Complex Space-Frequency Learning module, which performs unified complex-valued signal processing in both spatial and frequency domains, facilitating deep fusion through spatial structures and system-level characteristics. By leveraging the holistic representation capability of complex-valued neural networks, CompEvent achieves full-process spatiotemporal fusion, maximizes complementary learning between modalities, and significantly strengthens low-light video deblurring capability. Extensive experiments demonstrate that CompEvent outperforms SOTA methods in addressing this challenging task.
\end{abstract}

\begin{links}
    \link{Code}{https://github.com/YuXie1/CompEvent}
\end{links}

\section{Introduction}

\begin{figure}[t]
\centering
\includegraphics[width=\linewidth]{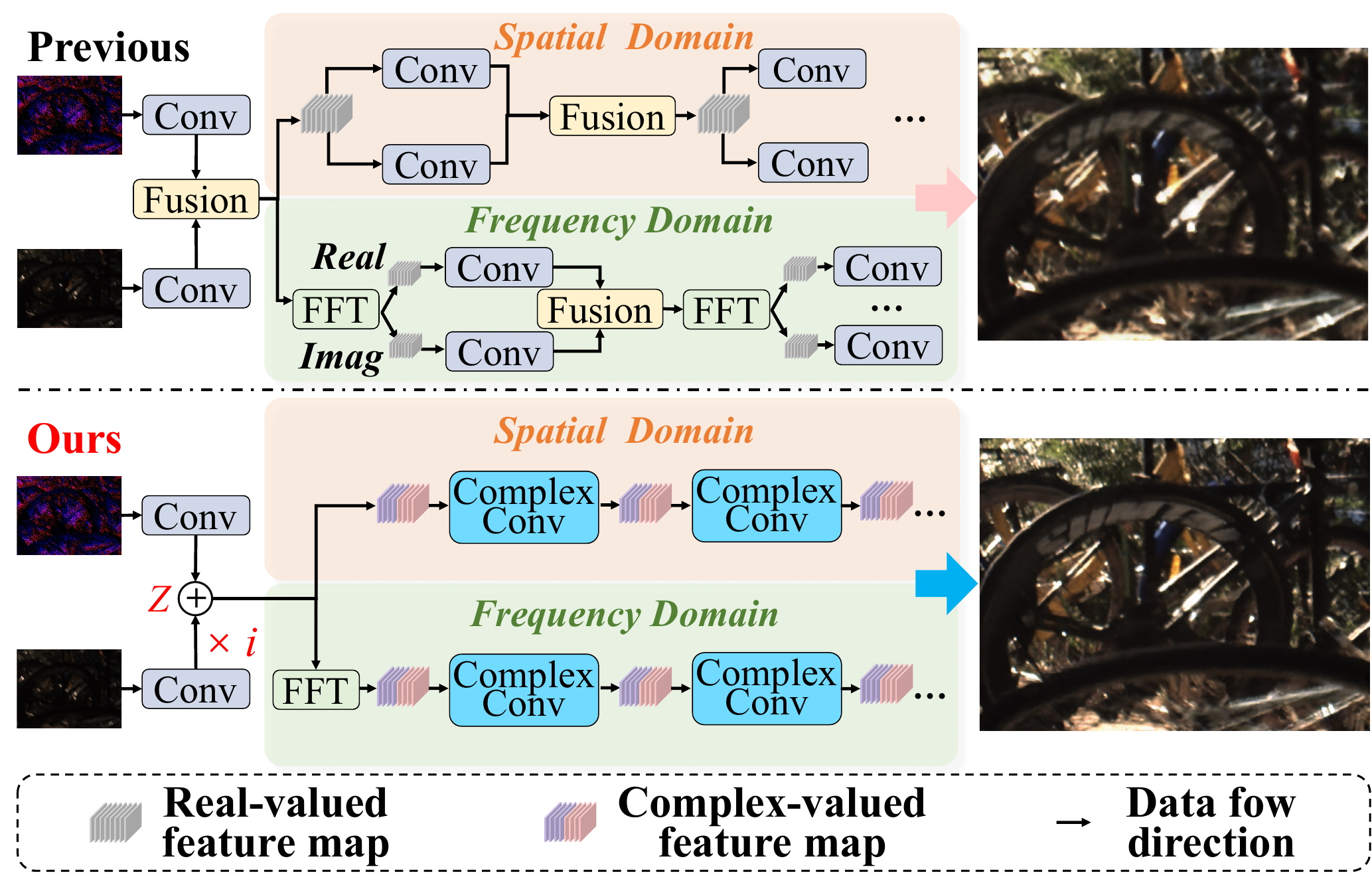}
\caption{Comparison of our method with previous methods. Previous methods perform fusion in a staged manner and split complex features into real-valued components before convolution in the frequency domain. Our method uses complex representations for both modalities, enabling full-process fusion by interacting features during processing. Moreover, our method directly applies complex convolutions to frequency features without separating them.}
\label{intro}
\vspace{-1em}
\end{figure}

In applications such as nighttime surveillance and autonomous driving, video capture in low-light environments inevitably requires extended exposure times, often suffering from the dual degradations of insufficient brightness and motion blur, leading to a sharp decline in video quality~\cite{kim2024towards}. These two degradations are tightly coupled: the long exposure required to increase brightness actually exacerbates the blur of moving objects and obliterates significant edge and texture details. This makes the joint task of video enhancement and deblurring a highly ill-posed problem.

Traditional video restoration methods primarily rely on information within the frame itself and face a fundamental information bottleneck when dealing with this compound degradation. Early model-driven approaches~\cite{horn1981determining,lucas1981iterative} and more recent deep learning-driven approaches, including convolutional neural networks (CNNs)~\cite{nah2017deep} and Transformers~\cite{vaswani2017attention,wang2022uformer,zamir2022restormer}, perform poorly in low-light scenarios. This is because, in low-light conditions, the extended exposure time required to compensate for brightness exacerbates motion blur and significantly loses edge and texture details. Compared to deblurring under normal exposure, models face a more challenging task in estimating the blurring process in low-light conditions. Simply combining low-light enhancement and deblurring tasks in tandem often leads to suboptimal results due to the accumulation and amplification of errors (e.g., noise is enhanced and blur is solidified)~\cite{zhou2022lednet}.

Event cameras, with their unique advantages of high temporal resolution (HTR) and high dynamic range (HDR), offer new possibilities for addressing this problem (Ruedi 1996; Lichtsteiner and Delbruck 2005)~\cite{lichtsteiner200364x64}. Their HTR features capture fine motion trajectories to guide deblurring, while their HDR features perceive scene structure in extremely dark environments to aid low-light enhancement~\cite{brandli2014240}. The precise spatiotemporal information provided by these two features perfectly complements the rich texture and color of RGB frames.

However, effectively fusing the advantages of these two heterogeneous modalities remains an open challenge. Existing event-RGB fusion methods, including some pioneering work~\cite{kim2024towards}, mostly follow a "staged fusion" strategy. In this paradigm, the network processes the two modalities in independent streams, exchanging information only at specific, discrete nodes, as shown in Figure~\ref{intro}. This discontinuous fusion approach fundamentally limits the network's ability to learn a deep, collaborative feature representation. Between the two fusion nodes, the network is forced to independently learn suboptimal single-modal features, making the fusion itself more of a "patching" operation than a deep integration. This approach fails to fully exploit the fine-grained spatiotemporal correlations between the two data sets, severely limiting its performance in scenarios that rely heavily on complementary information, such as intense motion or extreme dimming. Therefore, we believe that a more optimal solution should implement deep interaction throughout, allowing the features of the two modalities to co-evolve at every layer of the network processing, thereby building a comprehensive and robust understanding of degraded scenes.

To overcome the limitations of "staged fusion," we propose a novel restoration framework, CompEvent, to achieve "full-process fusion." The core idea of CompEvent is to leverage the inherent coupling properties of complex algebra to achieve deep interaction between modalities. Specifically, we unify the low-light blur RGB features and high-temporal-resolution event features as the real and imaginary parts of a complex tensor. To achieve this full-process fusion paradigm, CompEvent's architecture consists of two core complex-domain components. The first is the Complex Temporal Alignment Gated Recurrent Unit, which extends the GRU mechanism, known for its temporal processing capabilities, to the complex domain. It aligns and fuses video and event streams through complex convolutions, robustly processing temporal information in a recursive manner. The time-aligned features are then fed into the second core component, the Complex Space-Frequency Learning (CSFL) module. This module, serving as the backbone of the network, collaboratively performs spatial and frequency domain processing in a unified complex domain, achieving joint restoration by deeply integrating the spatial structure and frequency representations of the scene. Using complex operations to process the Fourier spectrum avoids the information fragmentation caused by the forced separation of real and imaginary components in traditional real networks (as shown in Figure~\ref{intro}). Leveraging the holistic representational power of complex networks, CompEvent internalizes modal fusion into its fundamental operations, enabling full-process spatiotemporal fusion and maximizing complementary learning between modalities. 

In summary, our contributions are as follows:
\begin{itemize}
\item We propose CompEvent, a complex-valued event-RGB video restoration framework that integrates modal fusion throughout the entire process of feature extraction, alignment, and restoration, fully leveraging the complementary advantages of event and RGB.
\item We design the Complex Temporal Alignment Gated Recurrent Unit, which organically combines the inherent fusion capabilities of complex operations with the temporal modeling advantages of recurrent neural networks, achieving temporal alignment that is robust to severely degraded videos.
\item We construct the Complex Spatial-Frequency Learning module, which synergistically processes spatial structure and frequency representations in the unified complex domain. This module can more effectively utilize the fused multimodal information to jointly correct motion blur and low-light effects.

\end{itemize}

Experiments on multiple benchmarks show that CompEvent outperforms state-of-the-art methods on the joint task of low-light video enhancement and deblurring, validating its effectiveness.

\section{Related Work}

\subsection{Motion Deblurring}
Traditional video deblurring relies on frames alone. Early methods were model-based~\cite{levin2009understanding}. With the rapid development of deep learning~\cite{li2023edge,li2024fouriermamba,li2024noise,jiang2024rbsformer,zhu2024learning,li2025enhanced,li2025learnable}, modern approaches leverage deep learning networks, from CNNs~\cite{nah2017deep} to Transformers~\cite{wang2022uformer,zamir2022restormer}. However, in low-light scenes, long exposure times worsen motion blur and severely degrade edge and texture detail~\cite{kim2024towards}, making motion estimation and detail recovery extremely difficult. 
To address this, researchers use event cameras, which asynchronously record brightness changes with high temporal resolution, capturing motion trajectories lost in blurred frames. Events thus serve as effective motion priors for handling severe blur~\cite{qi2024deblurring,liang2023coherent}.

\subsection{Low-light Enhancement}
Low-light enhancement methods are also predominantly frame-based. They include Retinex-based models~\cite{land1977retinex,wei2018deep} that decompose images into illumination and reflection, and zero-reference methods~\cite{guo2020zero,jiang2021enlightengan} that learn enhancement without ground truth. However, applying these methods directly to blurry low-light videos amplifies noise and artifacts~\cite{zhou2022lednet}.
Event cameras, due to their high dynamic range, can preserve scene structure even in underexposed areas. Introducing events provides structural priors absent in frames, aiding detail restoration and brightness improvement without blindly enhancing degradation~\cite{fu2024event,xu2024demosaicformer,liu2025event,sun2025evdm}.

\begin{figure*}[t]
\centering
\includegraphics[width=0.96\linewidth]{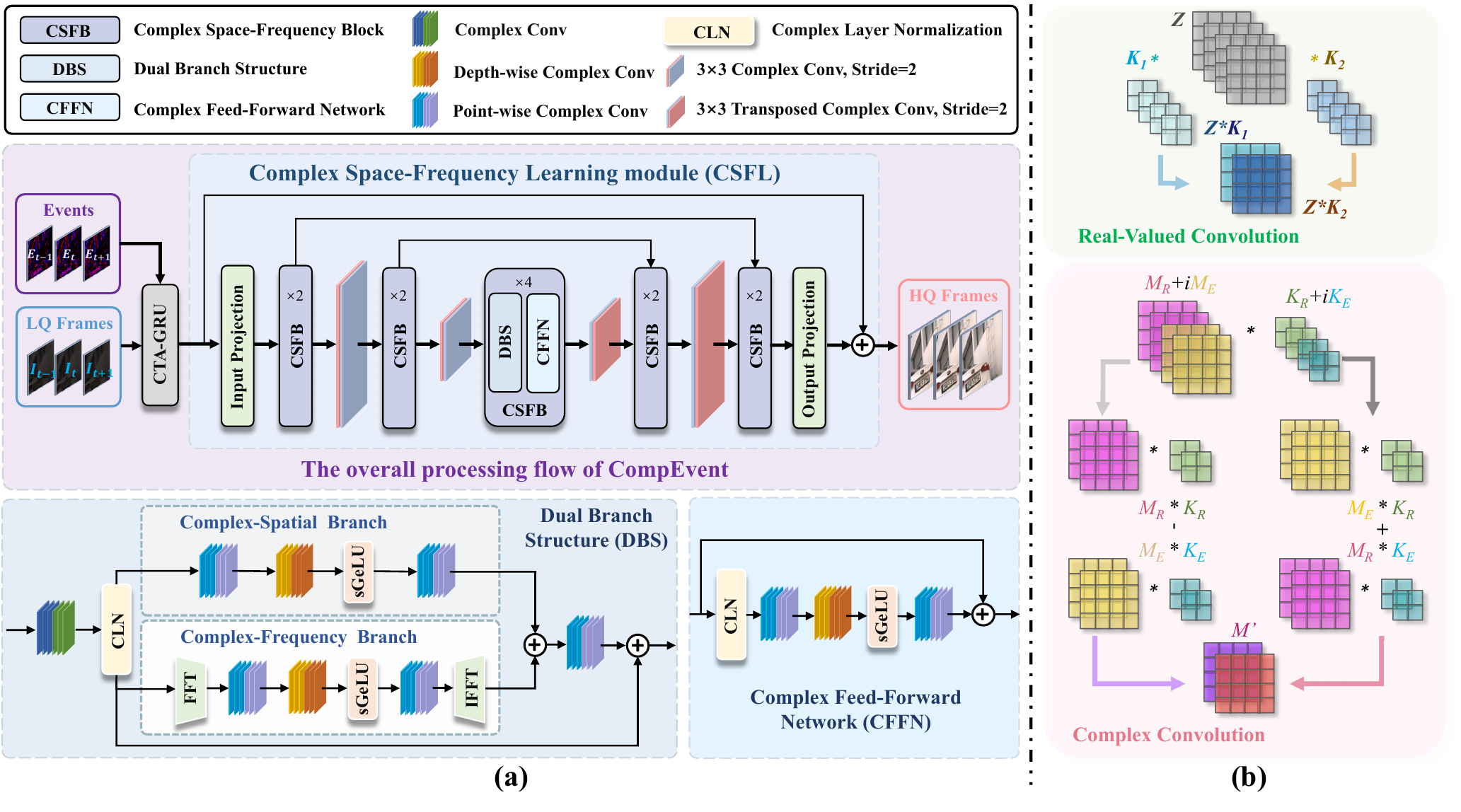}
\caption{(a) Overall architecture of the CompEvent framework. The Complex Temporal Alignment Gated Recurrent Unit (CTA-GRU) is shown in Figure~\ref{GRU}. (b) Comparison between the operations of complex convolution and real-valued convolution.}
\label{framework}
\vspace{-1em}
\end{figure*}

\subsection{Joint Low-light Enhancement and Deblurring}
Low-light and motion blur are physically coupled, motivating unified solutions. Frame-based methods such as LEDNet~\cite{zhou2022lednet} and JUDE~\cite{vo2025deep} improve upon cascaded models via joint architectures and deep algorithm but remain limited by frame information~\cite{kim2024towards}.
~\cite{kim2024towards} introduced the RELED dataset and ED-TFA, a staged event-frame fusion module. Yet staged fusion—common in prior work—restricts cross-modal interaction to specific stages, limiting recovery under extreme degradation. We instead propose a full-process fusion strategy using complex neural networks, enabling continuous spatiotemporal integration to more effectively solve this joint restoration task.

\begin{figure*}[h]
\centering
\includegraphics[width=0.95\linewidth]{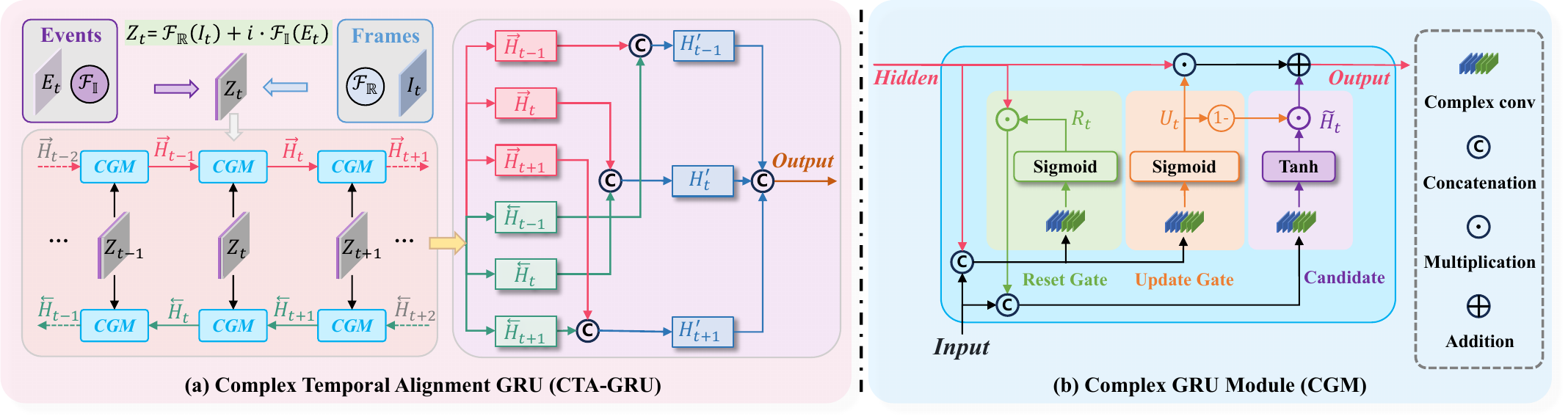}
\caption{Complex Temporal Alignment Gated Recurrent Unit (CTA-GRU). It consists of multiple cascaded Complex GRU Modules (CGM) with a bidirectional architecture, enabling the fusion of temporal information from both past and future frames.}
\label{GRU}
\vspace{-1em}
\end{figure*}

\section{Methodology}

\subsection{Overall Framework}
Figure~\ref{framework} illustrates the overall architecture of the proposed CompEvent framework. At each time step \( t \), the low-light blurry RGB frame \( I_t \in \mathbb{R}^{H \times W \times 3} \) and the corresponding event representation \( E_t \in \mathbb{R}^{H \times W \times C_E} \) are processed by two separate embedding networks: \( \mathcal{F}_\mathbb{R} \) and \( \mathcal{F}_\mathbb{I} \), each composed of several convolutional layers. These networks extract the real and imaginary components of the complex-valued representation, which are then combined into a complex tensor:
\[
Z_t = \mathcal{F}_\mathbb{R}(I_t) + i \cdot \mathcal{F}_\mathbb{I}(E_t)
\]
where \( i \) is the imaginary unit, and \( H, W, C \) represent the height, width, and number of channels, respectively. This representation is not a simple concatenation of the two modalities along the channel dimension. Instead, it leverages the complex convolution algorithm to inherently promote the joint learning of features from both the real and imaginary parts, thereby achieving more effective information fusion. In real-valued convolution, the convolution operation between the feature map \( Z \) and the kernel \( K \) is expressed as \( Z * K \), where \( * \) represents the convolution operation. In complex convolution, the convolution result M' , of the complex-valued feature maps \( M = M_R + iM_E \) and the complex kernel \( K = K_R + iK_E \) is:
\[
\begin{aligned}
M' &= K * M \\
&= \left(K_R * M_R - K_E * M_E\right) +
\\
&\quad i \cdot \left(K_R * M_E + K_E * M_R\right)
\end{aligned}
\]

As shown in the Figure~\ref{framework}, complex convolution jointly operates on the real part \( M_R \) and the imaginary part \( M_E \) via a shared kernel consisting of \( K_R \) and \( K_E \)~\cite{luo2025pancomplex}. This operation mechanism allows each complex output \( M' \) to depend on both modalities, achieving tighter fusion than real convolution. Furthermore, compared to real convolution, this shared structure reduces the number of parameters by nearly $50\%$ while enhancing cross-modal learning. Complex convolution naturally supports a "full-process fusion" strategy, promoting the continuous interaction of RGB and event features throughout the network.

The overall processing flow of CompEvent is shown in Figure~\ref{framework}: CompEvent receives three consecutive frames of complex features \( \{Z_{t-1}, Z_t, Z_{t+1}\} \). These are first input into the CTA-GRU module for temporal alignment. This module models temporal relationships in the complex-valued domain and robustly aligns the features by incorporating the context of the previous and next frames. Subsequently, the aligned features \( \{H'_{t-1}, H'_t, H'_{t+1}\} \) are concatenated and fed into the CSFL module. CSFL uses a hierarchical U-Net structure, combines spatial and frequency information modeling details, outputs a restored residual map for each frame, and adds it to the input image to ultimately generate a clear and bright video sequence.

\subsection{Complex Temporal Alignment GRU}
Accurate temporal alignment of features becomes highly challenging under severe motion blur and low-light noise. Traditional optical flow methods~\cite{horn1981determining,lucas1981iterative}, rely on the assumptions of constant brightness and spatial smoothness. However, these assumptions often fail under conditions of blur caused by long exposures and noise caused by low light, resulting in motion estimation failures.To address this, we adopt recurrent neural networks (RNNs) to model temporal relations implicitly, avoiding explicit motion estimation. Specifically, we employ gated recurrent units (GRUs) , which flexibly regulates information flow through a gating mechanism and can maintain higher stability in the face of uncertainty and noise~\cite{zhou2022lednet}.

To better utilize the high temporal resolution of event data, we extend GRU to the complex-valued domain and propose CTA-GRU. Let \( Z_t \in \mathbb{C}^{H \times W \times C} \) denote the complex-value input at time \( t \), and \( H_{t-1} \in \mathbb{C}^{H \times W \times C} \) the previous hidden state. The complex reset and update gates are computed as:
\[
R_{t}=\sigma_{c}\left(CConv_{r}\left(\left[Z_{t}, H_{t-1}\right]\right)\right)
\]
\[
U_{t}=\sigma_{c}\left(CConv_{u}\left(\left[Z_{t}, H_{t-1}\right]\right)\right)
\]
where \( CConv \) denotes complex convolution, \( \sigma_c \) is the complex sigmoid, and \( [\cdot, \cdot] \) denotes channel-wise concatenation. The complex candidate hidden state is:
\[
\tilde{H}_{t}=\tanh _{c}\left(CConv_{h}\left(\left[Z_{t}, R_{t} \odot H_{t-1}\right]\right)\right)
\]
and the updated hidden state becomes:
\[
H_{t}=\left(1-U_{t}\right) \odot H_{t-1}+U_{t} \odot \tilde{H}_{t}
\]
Where, \( \odot \) is the complex Hadamard product. We adopt split activations: for input \( z = x + iy \), we define \( \sigma_c(z) = \sigma(x) + i\sigma(y) \), which has been proven effective and stable in practice~\cite{nah2017deep,zamir2022restormer}.

The core advantage of the CTA-GRU lies in its gating mechanism being driven by complex convolutions, meaning that the reset and update gate decisions are based on a deep fusion of RGB features (real part) and event features (imaginary part). For example, when processing a fast-moving object, the event stream of the current frame (the imaginary part of \(Z_t\)) provides a clear motion trajectory. This information is passed to the reset gate \(R_t\) via the complex convolution \(CConv_r\). This gate "realizes" that the features corresponding to the object's old position in the previous hidden state \(H_{t-1}\) are outdated, and thus generates a smaller gate value to "reset" or ignore this information. 

To fully utilize the contextual information in a video, we use bidirectional CTA-GRU~\cite{wang2022uformer,jiang2021enlightengan}. A forward pass processes \( t-1 \to t \to t+1 \), yielding \( \{\overrightarrow{H}_{t-1}, \overrightarrow{H}_t, \overrightarrow{H}_{t+1}\} \), while a backward pass processes \( t+1 \to t \to t-1 \), yielding \( \{\overleftarrow{H}_{t+1}, \overleftarrow{H}_t, \overleftarrow{H}_{t-1}\} \). The aligned feature at \( t \) is:
\[
H_{t}'=concat\left(\overrightarrow{H}_{t}, \overleftarrow{H}_{t}\right)
\]
In this way, the features of each frame incorporate information from both the past and the future. Figure~\ref{GRU} illustrates the overall architecture of the proposed CTA-GRU. The aligned features \( \{H_{t-1}', H_t', H_{t+1}'\} \) are then concatenated and
fed into CSFL module for space-frequency restoration.

\subsection{Complex Space-Frequency Learning}
After temporal alignment through the CTA-GRU module, the features are fed into the backbone network for final image restoration. The mixed degradation of low light and motion blur exhibits different characteristics in different image domains: low light primarily affects low-frequency components (such as overall brightness and contrast)~\cite{huang2022deep,chen2023low}, while motion blur primarily manifests as an attenuation of high-frequency components (such as edges and texture)~\cite{manolakis2011applied}. Therefore, an ideal restoration network should be able to synergistically address the fine spatial structure and systematic frequency characteristics of the image.

\begin{figure*}
\centering
\includegraphics[width=0.85\linewidth]{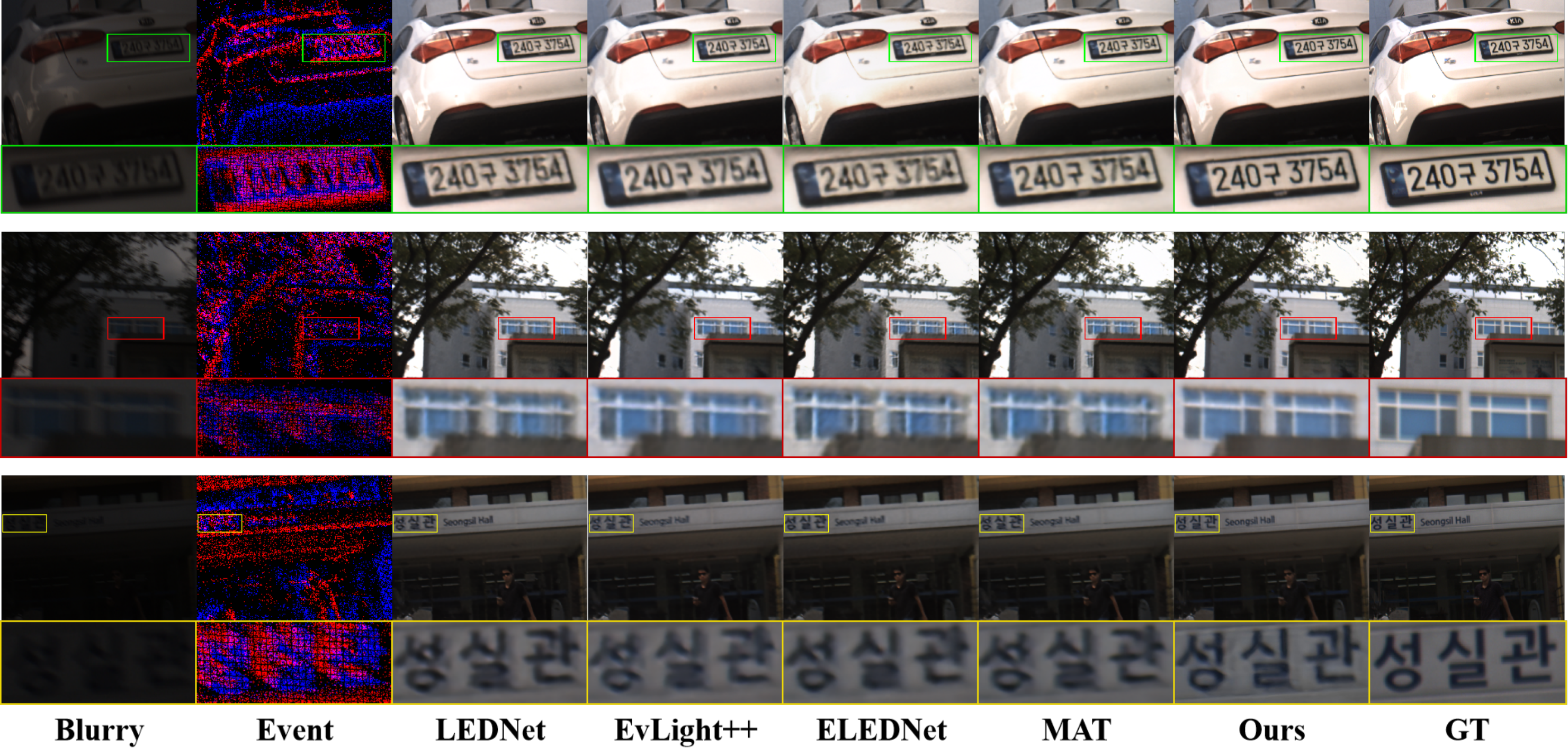}
\caption{Qualitative comparisons on the RELED dataset. Zoom in for better view.}
\label{RELED-visual}
\vspace{-1em}
\end{figure*}

The proposed Complex Space-Frequency Learning (CSFL) module adopts an encoder-decoder architecture shown in Figure~\ref{framework}, where downsampling and upsampling are handled by complex convolutions and transposed convolutions, respectively. Skip connections restore details at each level.Each Complex Space-Frequency Block (CSFB) processes input $X_{l-1} \in \mathbb{C}^{H \times W \times C}$ as follows. First, features are normalized via complex layer normalization (CLN) (See the supplementary material for details) :
\[
X_l' = \text{CLN}(X_{l - 1})
\]
We use whitening-based CLN~\cite{trabelsi2017deep} to preserve correlations between real and imaginary components, ensuring stability during training .

The features are then fed into a Dual Branch Structure (DBS)
. The first branch of DBS is the Complex Space Branch, which aims to learn spatially varying, deeply fused representations, extracting and fusing features simultaneously. Its main structure draws on depth-wise separable convolution~\cite{chollet2017xception,howard2017mobilenets}—an efficient form of convolution widely used in modern network architectures. We extend it to the complex domain to process the fused features. The operation can be expressed as:
\[
Y_{\text{spatial}} = \text{CConv}_{\text{pw2}} \bigl( \text{sGeLU} \bigl( \text{CConv}_{\text{dw}} \bigl( \text{CConv}_{\text{pw1}} (X') \bigr) \bigr) \bigr)
\]
Where, $\text{CConv}_{\text{pw1}}$ and $\text{CConv}_{\text{pw2}}$ are $1 \times 1$ complex point-wise convolutions used for channel mapping and mixing; $\text{CConv}_{\text{dw}}$ is a complex depth-wise separable convolution used to efficiently extract spatial features. Based on the complex activation function of \text{GeLU} , the \text{sGeLU} also employs a split activation strategy, applying the real-valued \text{GeLU} function to the real and imaginary parts of the complex input separately.

Another parallel branch of DBS is the Complex Frequency Branch, which processes systematic degradations. Features are first transformed via a two-dimensional fast Fourier transform (FFT) :
\[
\mathcal{F}(X_l') = \text{FFT2D}(X_l')
\]
The resulting complex spectrum $\mathcal{F}(X_l')$ is then processed in a complex convolutional network with a similar spatial branching structure :
\[
\mathcal{F}_{\text{proc}} = \text{CConv}_{\text{pw2}} \bigl( \text{sGeLU} \bigl( \text{CConv}_{\text{dw}} \bigl( \text{CConv}_{\text{pw1}} \bigl( \mathcal{F}(X_l') \bigr) \bigr) \bigr) \bigr)
\]
Then, the processed spectrum is converted back to the spatial domain via an inverse Fourier transform (IFFT) :
\[
Y_{\text{freq}} = \text{IFFT2D}(\mathcal{F}_{\text{proc}})
\]
The entire spectrum is processed holistically without separating real and imaginary parts, preserving the spectral structure and enabling adaptive corrections in the complex domain.

Outputs from both branches are combined with the input via a residual connection:
\[
X_l'' = X_{l - 1} + Y_{\text{spatial}} + Y_{\text{freq}}
\]

Finally, $X_l''$ passes through a Complex Feed-Forward Network (CFFN) for further refinement:
\[
X_l = \text{CFFN}\bigl(\text{CLN}(X_l'')\bigr) + X_l''
\]
Where, $\text{CFFN}(X) = \text{CConv}_{\text{pw4}} \bigl( \text{sGeLU} \bigl( \text{CConv}_{\text{pw3}}(X) \bigr) \bigr)$, $\text{CConv}_{\text{pw3}}$ and $\text{CConv}_{\text{pw4}}$ are $1 \times 1$ complex point-wise convolutions. CFFN performs nonlinear transformations directly in the complex domain. It not only models complex feature relationships like real-valued FFN, but also preserves and utilizes the phase information of the signal, thus possessing richer representation capabilities~\cite{bassey2021survey}. By stacking multiple such CSFBs, our network can deeply and collaboratively process spatial and frequency domain information at different scales, thereby achieving effective restoration of low-light blurry videos.

\begin{table*}[tb]
\centering
\begin{tabular*}{\textwidth}{@{\extracolsep{\fill}}
>{\centering\arraybackslash}m{3.2cm}|
>{\centering\arraybackslash}m{6cm}|
>{\centering\arraybackslash}m{1cm}|
>{\centering\arraybackslash}m{1.2cm}
>{\centering\arraybackslash}m{1.2cm}|
>{\centering\arraybackslash}m{1.2cm}
>{\centering\arraybackslash}m{1.2cm}}
\toprule
\multicolumn{2}{c|}{\multirow{2}{*}{\textbf{Methods}}}  & \multirow{2}{*}{\textbf{Input}}  & \multicolumn{2}{c|}{\textbf{RELED}} & \multicolumn{2}{c}{\textbf{LOL-Blur}} \\
\multicolumn{2}{c|}{} & & \textbf{PSNR} & \textbf{SSIM} & \textbf{PSNR} & \textbf{SSIM} \\
\hline
\multirow{5}{*}{\centering\arraybackslash \makecell[c]{Low-Light \\ Enhancement}}
  & SNRNet~\cite{xu2022snr}         & F & 26.47 & 0.851 & 20.25 & 0.815 \\
  & LLFormer~\cite{wang2023ultra}       & F & 26.62 & 0.862 & 20.68 & 0.832 \\
  & RetinexFormer~\cite{cai2023retinexformer}  & F & 26.66 & 0.865 & 20.83 & 0.817 \\
  & SDSDNet~\cite{wang2021seeing}       & F & 28.47 & 0.887 & 21.34 & 0.832 \\
  & EvLight++~\cite{chen2024evlight++}     & F+E & 30.87 & 0.888 & 24.99 & 0.880 \\
\hline
\multirow{11}{*}{Motion Deblur}
  & MPRNet~\cite{zamir2021multi}         & F   & 26.89 & 0.867 & 21.35 & 0.825 \\
  & MIMOUNet+~\cite{cho2021rethinking}      & F   & 26.52 & 0.866 & 21.12 & 0.821 \\
  & NAFNet~\cite{chen2022simple}         & F   & 26.77 & 0.862 & 21.28 & 0.818 \\
  & RNN-MBP~\cite{zhu2022deep}       & F   & 29.52 & 0.902 & 23.58 & 0.862 \\
  & DSTNet~\cite{pan2023deep}       & F   & 29.59 & 0.903 & 23.63 & 0.864 \\
  & e-SLNet~\cite{wang2020event}       & F+E & 19.45 & 0.663 & 17.05 & 0.738 \\
  & REDNet~\cite{xu2021motion}        & F+E & 29.19 & 0.903 & 23.25 & 0.859 \\
  & EFNet~\cite{sun2022event}          & F+E & 29.85 & 0.905 & 23.92 & 0.867 \\
  & MAT~\cite{xu2025motion}           & F+E & 31.22 & 0.896 & 25.15 & \underline{0.882} \\
  & UEVD~\cite{kim2022event}       & F+E & 29.93 & 0.905 & 24.08 & 0.869 \\
  & REFID~\cite{sun2023event}     & F+E & 30.10 & 0.913 & 24.55 & 0.875 \\
\hline
\multirow{1}{*}{Joint (Frame-based)}
  & LEDNet~\cite{zhou2022lednet}         & F & 30.36 & 0.887 & \underline{25.74} & 0.850 \\
\hline
\multirow{2}{*}{Joint (Event-guided)}
  & ELEDNet~\cite{kim2024towards}        & F+E & \underline{31.30} & \underline{0.925} & 25.04 & 0.873 \\
  & \textbf{Ours} & F+E & \textbf{32.51} & \textbf{0.928} & \textbf{28.73} & \textbf{0.907} \\
\bottomrule
\end{tabular*}
\caption{The quantitative results on RELED and LOL-Blur. “F” denotes image frame-based methods, while “F+E” represents frame-based methods integrated with event-guided information. Best and second-best results are boldfaced and underlined.}
\label{Baseline}
\vspace{-1em}
\end{table*}

\section{Experiments and Analysis}
\subsection{Datasets}
We evaluate CompEvent on the RELED real-world dataset~\cite{kim2024towards} and the LOL-Blur synthetic dataset~\cite{zhou2022lednet}. Training details and hyperparameters are provided in the supplementary material.
(1) \textbf{RELED}~\cite{kim2024towards} is the first large-scale real-world benchmark built for the joint low-light enhancement and motion deblurring tasks. The dataset is acquired through an optical beam splitting system that can simultaneously record low-light blurry videos, the corresponding high-quality clear images, and high-fidelity event streams. (2) \textbf{LOL-Blur}~\cite{zhou2022lednet} is a large-scale synthetic dataset that provides low-light blurry image and clear image pairs for the joint low-light enhancement and deblurring tasks. To adapt our event-based approach, we generate the corresponding event streams for it using the ESIM simulator~\cite{rebecq2018esim}.

\subsection{Comparison with State-of-the-Art Methods}
To comprehensively evaluate the CompEvent framework, we conduct extensive comparisons against state-of-the-art methods across a variety of tasks and modalities. These baselines are systematically categorized into four categories: single low-light enhancement methods, single motion deblurring methods, frame-only joint restoration methods, and event-guided joint restoration methods. The detailed description of the experimental setup for the baseline comparison method is provided in the supplementary material.

\begin{figure*}
\centering
\includegraphics[width=0.85\linewidth]{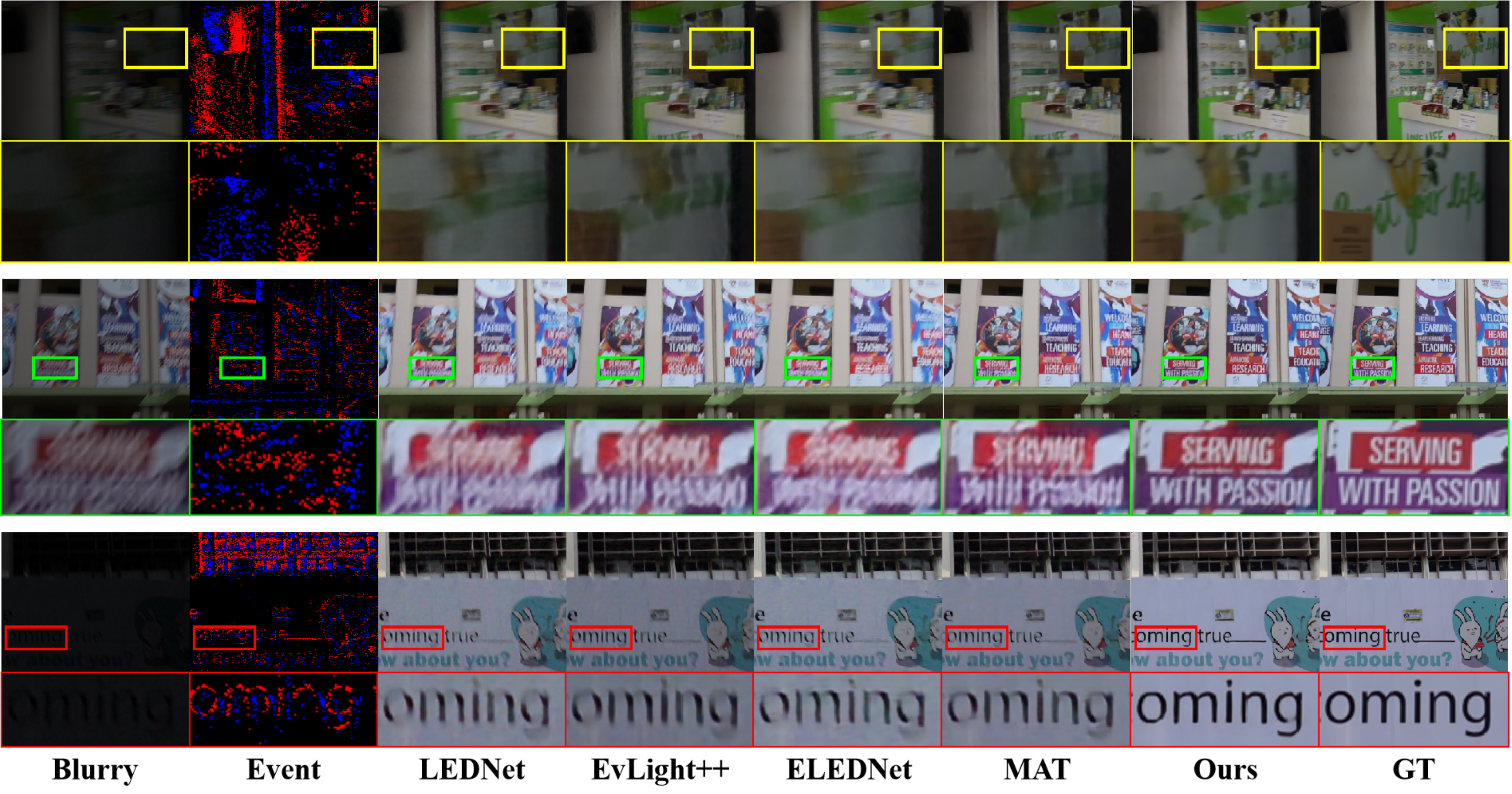}
\caption{Qualitative comparisons on the LOL dataset. Zoom in for better view.}
\label{LOL-visual}
\vspace{-1.5em}
\end{figure*}

\subsubsection{Real-World Dataset (RELED):}
As shown in Table~\ref{Baseline}, CompEvent achieves a PSNR of 32.51 dB and an SSIM of 0.928 on the real-world RELED dataset, outperforming all single-task methods, including EvLight++ (30.87 / 0.888) for low-light enhancement and MAT (31.22 / 0.896) for deblurring, as well as joint frameworks such as LEDNet (30.36 / 0.887) and ELEDNet (31.30 / 0.925). As shown in Figure~\ref{RELED-visual}, CompEvent produces sharper structures with fewer artifacts under complex lighting and motion conditions. These results demonstrate the effectiveness of our method for real-world video restoration.

\subsubsection{Synthetic Dataset (LOL-Blur):}
As shown in Table~\ref{Baseline}, CompEvent achieves 28.73 / 0.907 PSNR/SSIM, surpassing the best single-task baselines EvLight++ (24.99 / 0.880) and MAT (25.15 / 0.882) by large margins. Compared to joint methods LEDNet (25.74 / 0.850) and ELEDNet (25.04 / 0.873), CompEvent improves by up to +3.69 dB and +0.057 SSIM. Qualitative results in Figure~\ref{LOL-visual} confirm its superior texture and structural recovery under synthetic degradation.

\subsection{Ablation Studies}
We conduct ablation studies on the RELED dataset to evaluate the effectiveness of each component in our CompEvent framework. By systematically removing or substituting key modules, we quantify their contributions.

\begin{table}[tb]
\centering
\small
\begin{tabular}{l|cc}
\toprule
\textbf{Model Variant} & \textbf{PSNR} & \textbf{SSIM} \\
\hline
(a) CompEvent (Full) & 32.51 & 0.928 \\
(b) w/o Complex (Concat) & 31.39 \textbf{(-1.12↓)} & 0.901 \textbf{(-0.027↓)} \\
(c) w/o GRU (Static) & 30.87 \textbf{(-1.64↓)} & 0.885 \textbf{(-0.043↓)} \\
(d) w/o GRU (Concat) & 31.93 \textbf{(-0.58↓)} & 0.914 \textbf{(-0.014↓)} \\
(e) w/o Freq. Branch & 31.76 \textbf{(-0.75↓)} & 0.908 \textbf{(-0.020↓)} \\
\bottomrule
\end{tabular}
\caption{Ablation study of the core components of CompEvent on the RELED dataset.}
\label{abl_structure}
\vspace{-1.5em}
\end{table}

\subsubsection{Effectiveness of Complex Full-Process Fusion.}
To evaluate our fusion design, we construct a real-valued variant (b) without Complex (Concat), replacing complex convolution with a real-valued counterpart of similar parameter size, where RGB and event features are concatenated in the channel dimension. Results in Table~\ref{abl_structure} show that this variant achieves a 1.12 dB decrease in PSNR, demonstrating that the performance improvement does not come from more parameters, but rather benefits from the inherent fusion mechanism brought by complex convolution throughout the network, which enables the network to more effectively utilize the complementary advantages of the two modalities to cope with complex mixed degradations.

\subsubsection{Effectiveness of CTA-GRU for Temporal Modeling.}
We designed two variants to evaluate the contribution of CTA-GRU: (c) without GRU (Static), completely removing the GRU; and (d) without GRU (Concat), replacing the GRU model with cross-frame concatenation. Results in Table~\ref{abl_structure} show that removing the CTA-GRU significantly degrades model performance, demonstrating the critical importance of modeling temporal context for video restoration. Furthermore, compared to cross-frame concatenation, CTA-GRU achieves dynamic alignment between frames through a gating mechanism, demonstrating greater robustness to large motion and inter-frame misalignment.

\subsubsection{Effectiveness of CSFL Module.}
We validated the design of the Complex Space-Frequency Learning (CSFL) module. This module consists of a spatial branch and a frequency branch. To evaluate the contribution of the frequency branch, we constructed a variant (e) w/o Freq. Branch, which removes the frequency branch while retaining the spatial branch. Results in Table~\ref{abl_structure} show that this variant significantly degrades performance, validating that separately modeling blur (high-frequency) and low-light (low-frequency) degradation in the frequency domain is an effective strategy for addressing this mixed degradation problem. The complete CSFL module achieves the best overall performance through space-frequency co-processing.

\subsubsection{Effectiveness of Complex Layer Normalization (CLN).}
To evaluate CLN, we test (f) w/o CLN (Separate), applying real-valued normalization to real and imaginary parts separately, and (g) w/o CLN (Concat), concatenating and normalizing them jointly. Results in Table~\ref{abl_norm} show that both alternatives lead to performance degradation, with (g) performing the worst. This suggests that directly treating complex features as ordinary real-valued vectors for normalization destroys their algebraic structure and weakens their expressive power. In contrast, CLN normalizes the variance of the real and imaginary parts by using a whitening transformation and considers their covariance to decorrelate them, improving training stability and final performance.

\begin{table}[tb]
\centering
\small  
\begin{tabular}{l|cc}
\toprule
\textbf{Model Variant} & \textbf{PSNR} & \textbf{SSIM} \\
\hline
(a) CompEvent (Full) & 32.51 & 0.928 \\
(f) w/o CLN (Separate) & 31.98 \textbf{(-0.53↓)} & 0.915 \textbf{(-0.013↓)} \\
(g) w/o CLN (Concat) & 31.55 \textbf{(-0.96↓)} & 0.904 \textbf{(-0.024↓)} \\
\bottomrule
\end{tabular}
\caption{Ablation study of the Complex Layer Normalization (CLN) on the RELED dataset.}
\label{abl_norm}
\vspace{-1.5em}
\end{table}

\section{Conclusion}
We propose CompEvent, a complex-valued neural network framework for joint low-light video enhancement and deblurring, enabling holistic full-process fusion of event data and RGB frames. It features two core components: the Complex Temporal Alignment GRU for efficient temporal alignment and recursive fusion via complex operations, and the Complex Space-Frequency Learning module for synergistic spatial and frequency domain processing in a unified complex domain. Throughout the process, CompEvent overcomes "staged fusion" limitations
by leveraging the inherent complex-valued operations of complex convolution to fuse RGB and event information at every step, enabling full-process spatiotemporal fusion.
Extensive experimental results on several benchmarks demonstrate the effectiveness of the proposed method.

\section{Acknowledgments}
This work was supported by the National Natural Science Foundation of China (NSFC) under Grants 62225207, 62436008, 62422609 and 62276243.

\bibliography{aaai2026}

\end{document}